\documentclass{article}

\PassOptionsToPackage{numbers, compress}{natbib}

\usepackage[final]{main}

\usepackage[utf8]{inputenc} 
\usepackage[T1]{fontenc}    
\usepackage{hyperref}       
\usepackage{url}            
\usepackage{booktabs}       
\usepackage{amsfonts}       
\usepackage{nicefrac}       
\usepackage{microtype}      

\usepackage{amsmath}
\usepackage{amssymb}
\usepackage{graphicx}       
\usepackage{arydshln}

\title{Spatio-Temporal Graph Convolutional Networks: Optimised Temporal Architecture}

\author{%
    Edward Turner\thanks{This work was done as an MMath student within the Department of Statistics at the University of Oxford.}\\
    Department of Statistics\\
	University of Oxford\\
    \href{mailto:edward.turner01@outlook.com}{\texttt{edward.turner01@outlook.com}}\\
}

\begin{document}

\maketitle

\begin{abstract}
Spatio-Temporal graph convolutional networks were originally introduced with CNNs as temporal blocks for feature extraction. Since then LSTM temporal blocks have been proposed and shown to have promising results. We propose a novel architecture combining both CNN and LSTM temporal blocks and then provide an empirical comparison between our new and the pre-existing models. We provide theoretical arguments for the different temporal blocks and use a multitude of tests across different datasets to assess our hypotheses.
\end{abstract}

\section{Introduction}
The field of graph representation learning (GRL) aims to take a feature rich graph (e.g. a social network with edges representing friendships and vertices holding information on each individual) and then in a low-dimensional vector space, the latent space, produce a feature representation embedding of the graph, the latent representation \cite{turner2021graph}. This provides a better environment for inference of the data where commonly one would be tasked with vertex prediction, the prediction of vertex labels for some or all of the vertices, or relation prediction, the prediction of whether two nodes are linked by an edge or not \cite{grl}. In this paper we primarily focus on vertex prediction.

Spatio-temporal graph convolutional networks (ST-GCNs) are a specific class of GRL methods that arose in $2018$ from the separate works of Yan et al. and Yu et al. \cite{Yan_2018, Yu_2018}. The developed frameworks both attempt to model time-series node features while having a static graph structure for node interaction. They achieve this by stacking convolutional neural networks (CNNs), which the vertex features are passed through to convolve the time-series data, with graph convolutional networks (GCNs), which the graph with updated convolved features is passed through. 

\paragraph{Our Contributions} We investigate the use of CNN versus long short-term memory (LSTM) models for the temporal blocks in a ST-GCN model. We also look at a combination model including both temporal block types, which, to the best of our knowledge, is a novel architecture. We then extensively compare models across a host of datasets ranging in size, complexity and noise.

\section{Preliminaries}
\subsection{Notation}
We consider a vertex prediction problem for an input graph $\mathcal{G} = \left(\mathcal{V}, \mathcal{E}, \mathcal{W}, \mathbf{X}\right)$, where $\mathcal{V}$ is the vertex set with $|\mathcal{V}|=N$, $\mathcal{E}$ is the edge set, $\mathcal{W}\in\mathbb{R}^{N\times{N}}$ is the weighted adjacency matrix and $\mathbf{X}\in\mathbb{R}^{N\times{T}}$ is the vertex feature matrix, with each vertex feature recorded across T time intervals. From $\mathbf{X}$ a vertex feature tensor, $\mathcal{X}\in\mathbb{R}^{N\times{C_i}\times{T}}$ is created by convolving the matrix features into $C_i$ channels.

\subsection{Graph neural networks}
Graph neural networks (GNNs) overcome the challenge of permutation invariant deep learning on graph data via a message passing framework with multiple layers, similar to that of a classic neural network \cite{grl}. For an input graph $\mathcal{G} = \left(\mathcal{V}, \mathcal{E}, \mathcal{W}, \mathcal{X}\right)$, a GNN consists of an initial layer and then L further layers where the k\textsuperscript{th} layer contains $N$ nodes and may be represented as a tensor, $\mathbf{H}^{(k)} = \left(\mathbf{h}^{(k)}_1, ..., \mathbf{h}^{(k)}_N\right)^T\in\mathbb{R}^{N\times{C_k}\times{T}}$. Here $\mathbf{h}^{(k)}_i$ represents the i\textsuperscript{th} node in the k\textsuperscript{th} layer which corresponds to the i\textsuperscript{th} vertex of $\mathcal{G}$. A (self-loop) approach GNN is initialised by the following iteration:

\begin{equation}
    \mathbf{H}^{(0)} = \mathcal{X}\, ,
\end{equation}
\begin{equation}
\label{gnnequation}
    \mathbf{h}^{(k+1)}_i = \mathrm{Aggregate}^{(k)}({\{\mathbf{h}^{(k)}_i}\} \cup \{\mathbf{h}^{(k)}_j : e_{i, j} \in \mathcal{E}\}) \,, \hspace{2mm} 0 \le k \le L-1, \hspace{1mm} 1 \le i \le N,
\end{equation}

where the $\mathrm{Aggregate}$ functions are differentiable and may depend on $\mathcal{W}$ \cite{grl}.\footnote{The self-loop approach alleviates overfitting in the more general framework which has $\mathrm{Update}^{(k)}$ and $\mathrm{Aggregate}^{(k)}$ functions by combining the pair into a single $\mathrm{Aggregate}^{(k)}$ function \cite{grl}.} The first layer is set to be the vertex features and then for each following layer a vertex's embedding is based on its previous embedding updated with an aggregation of its neighbouring vertices' previous embeddings. We then take $\mathbf{H}^{(L)}$ to be our final low-dimension embedding of the graph (the latent representation).

\subsubsection{Graph convolutional networks}
GCNs take inspiration for their $\mathrm{Aggregate}$ functions from Fourier transformations in the spectral domain \cite{Yu_2018}. GCNs have seen many recent applications as in $2016$ Defferrard et al. provided graph convolution methods using a layer-wise propagation rule which has computational complexity $\mathcal{O}(N)$ rather than $\mathcal{O}(N^2)$ \cite{fast_local_spectral}. Kipf et al. subsequently simplified the GCN architecture, motivated by a first-order approximation of spectral graph convolutions \cite{DBLP:journals/corr/KipfW16}. Their model is able to encode both the local graph structure and the vertex features \cite{DBLP:journals/corr/KipfW16, kipf2016variational}. In a number of experiments on citation networks and a knowledge dataset, it was shown that the model outperforms recent related methods such as DeepWalk and Planetoid \cite{DBLP:journals/corr/KipfW16}. Defferrard et al. introduced the generalised K\textsuperscript{th} order Chebyshev convolution method, where a graph convolutional operator, $*_{\mathcal{G}}$, is used \cite{fast_local_spectral}. Let us define the normalised graph Laplacian $\mathcal{L}:=I_N - \mathcal{D}^{-\frac{1}{2}}\mathcal{W}\mathcal{D}^{-\frac{1}{2}}$, where $\mathcal{D}\in\mathbb{R}^{N\times{N}}$ is the diagonal matrix with $\mathcal{D}_{ii}=\Sigma_j \mathcal{W}_{ij}$, so that $\mathcal{D}^{-\frac{1}{2}}\mathcal{W}\mathcal{D}^{-\frac{1}{2}}$ is the symmetric normalisation of $\mathcal{W}$. By spectral decomposition $\mathcal{L}=U \Lambda U^T$, where $\Lambda \in \mathbb{R}^{N\times{N}}$ is the diagonal matrix of eigenvalues of $\mathcal{L}$ and $U\in\mathbb{R}^{N\times{N}}$ is the matrix of eigenvectors of $\mathcal{L}$, known as the graph Fourier basis \cite{signal_processing}. First we define $\Theta *_\mathcal{G} x := \Theta(\mathcal{L})x=\Theta(U\Lambda U^T)x=U\Theta(\Lambda)U^Tx$, for a singular graph signal $x \in\mathbb{R}^N$. Here the filter $\Theta(\Lambda)$ is also a diagonal matrix. By this definition, a graph signal $x$ is filtered by a kernel $\Theta$ with multiplication between $\Theta$ and $U^Tx$, the graph Fourier transform \cite{signal_processing}. 

To define $\Theta *_\mathcal{G} \mathcal{X} \in\mathbb{R}^{N\times{C_o}\times{T}}$, the output of the GCN, the vertex feature tensor, $\mathcal{X}$, is first decomposed into one signal per time dimension, $X^{(t)} = \left(x_1^{(t)}, ..., {{x_C}_h}^{(t)}\right) \in\mathbb{R}^{N\times{C_h}}$. Each are then passed in parallel through the GCN with a shared kernel, $\Theta\in\mathbb{R}^{N\times{C_h}\times{C_o}}$, according to:

\begin{equation}
    \left(\Theta *_\mathcal{G} X^{(t)}\right)_j := \sum_{i=1}^{C_h} \Theta^{(ij)}(\mathcal{L})x_i^{(t)} \,, \hspace{2mm} 1 \le j \le C_o,
\end{equation}

where $\left(\Theta *_\mathcal{G} X^{(t)}\right)_j$ is the j\textsuperscript{th} column of $\Theta *_\mathcal{G} X^{(t)} \in\mathbb{R}^{N\times{C_o}}$ \cite{Yu_2018}.

Similar to classic CNNs to localise the filter, reduce the number of parameters and reduce computational complexity Hammond et al. now proposed truncating the Chebyshev polynomial expansion of $\Theta(\Lambda)$ up to the K\textsuperscript{th} order \cite{wavelets}. This results in:

\begin{equation}
\label{eq:chebapprox}
  \left(\Theta *_\mathcal{G} X^{(t)}\right)_j \approx \sum_{k=0}^{K} \sum_{i=1}^{C_h} \Theta_k^{(ij)}T_k(\tilde{\mathcal{L}})x_i^{(t)} \, ,
\end{equation}

with a rescaled $\tilde{\mathcal{L}}=\frac{2}{\lambda_{\text{max}}}\mathcal{L}-I_N$ where $\lambda_{\text{max}}$ denotes the largest eigenvalue of $\mathcal{L}$ and $\Theta_k^{(ij)}$ is a matrix of trainable Chebyshev coefficients \cite{cheb_revisited, Yu_2018}. The Chebyshev polynomials are recursively defined as $T_0(x)=1$, $T_1(x)=x$ and $T_k(x) = 2xT_{k-1}(x) - T_{k-2}(x) \hspace{1.5mm} \forall k \geq 2$ \cite{wavelets}. Note, Equation~\hyperref[eq:chebapprox]{4} is now K-localized since it is a K\textsuperscript{th}-order polynomial in the Laplacian, that is it only depends on vertices that are at maximum K steps away from the central vertex.

In the case of K $=1$ Equation~\hyperref[eq:chebapprox]{4} reduces to the GCN used by Kipf et al. \cite{DBLP:journals/corr/KipfW16, kipf2016variational}. Letting $\tilde{\mathcal{W}}=\mathcal{W}+I_N$ and $\tilde{\mathcal{D}}_{ii}=\Sigma_j \tilde{\mathcal{W}}_{ij}$ this can be succinctly written as:\, $\Theta *_\mathcal{G} X^{(t)} = \Theta_1 (\tilde{D}^{-\frac{1}{2}}\tilde{W}\tilde{D}^{-\frac{1}{2}})X^{(t)}$, where $\Theta_1$ is now the trainable weight matrix \cite{DBLP:journals/corr/KipfW16}. Iteratively applying K first-order approximations achieves a similar effect as one passing of the K-order convolution, both methods drawing information from the K-step neighbourhood of each vertex. The layer-wise linear structure is parameter-economic and highly efficient for large-scale graphs, since the order of the approximation is limited to one \cite{Yu_2018}.

\subsection{Spatio-temporal graph convolutional networks}
\begin{figure}[t]
\centering
\includegraphics[scale=0.25]{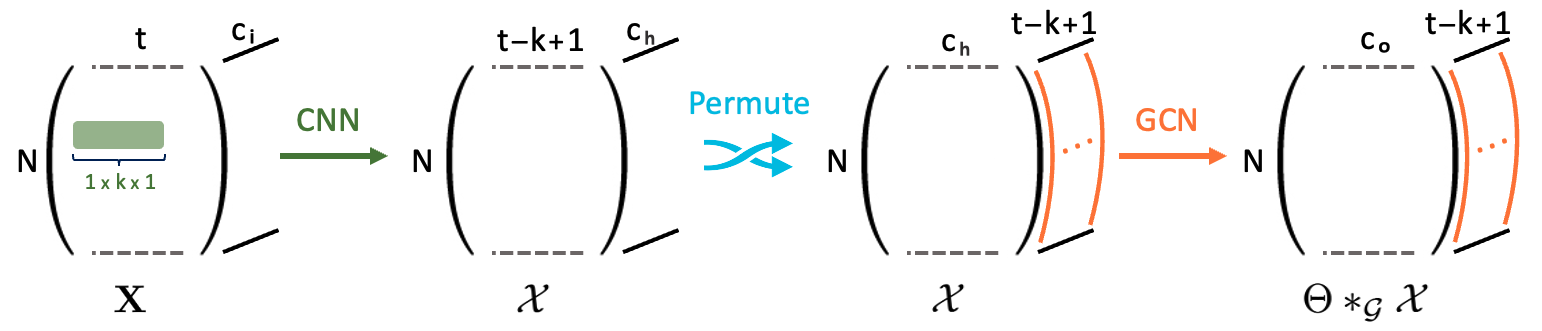}
\caption{\label{fig:framework}An overview of the message passing framework used by Yu et al. in their ST-GCN model \cite{Yu_2018}. In most input cases $C_i$ = 1, thus the $C_h$ channels are produced from different filters in the CNN. The tensor is then permuted with the channels becoming the vertex features for the GCN with each of the t-k+1 layers passed through in parallel.}
\end{figure}

ST-GCNs are formed of temporal blocks to handle the time-series vertex features and GCN blocks to pass the graph with newly convolved features. While Yan et al. handcraft their convolutional channels for $\mathcal{X}$, Yu et al. choose to use CNNs as their temporal block which we view as more general and thus focus on their framework further \cite{Yan_2018, Yu_2018}. Figure~\ref{fig:framework} shows the interaction of the temporal and spatial blocks used by Yu et al. \cite{Yu_2018}. Once $\Theta *_\mathcal{G} \mathcal{X}$ is obtained as described in Figure~\ref{fig:framework} Yu et al. then pass it through another CNN, repeat the whole process (to obtain a 2-step neighbourhood of information sharing) and then apply a linear layer for the final output \cite{Yu_2018}.

Others such as Simeunović et al. and Khodayar et al. use a LSTM temporal block instead of CNNs \cite{stgcn_power_forecasting, stgcn_wind}. We focus on block choice and architecture design for the remainder of this paper, discussing theoretical arguments for both block types in Section~\hyperref[sec:theory]{3} and then comparing performance in Section~\hyperref[sec:experiments]{4}.

\subsection{Finance dataset}
The finance dataset we use is a graph where the vertex set corresponds to $72$ of the largest S\&P\,$500$ companies. Using natural language processing (NLP) techniques the $72$ companies are detected from financial news articles on Reuters across $2007$ with entities that co-reference the same company, i.e. \textit{AAPL} and \textit{Apple}, merged. From this a news coverage matrix, $\mathbf{M}=(m_{i,j})\in\mathbb{R}^{72\times{12,311}}$, is constructed where each row represents a company and each column represents one of the $12,311$ news articles, with $m_{i,j} := \delta_{\mathrm{company}\, i\, \in\, \mathrm{news}\, j}$. We then take the edge weight between any two vertices to be the cosine distance of their rows in  $\mathbf{M}$. The associated vertex features are the daily returns for the $2007$ financial year, consisting of $250$ trading days. The graph form is a PyTGT StaticGraphTemporalSignal dataset and thus can be used with all relevant models in the package \cite{rozemberczki2021pytorch}.

\section{Temporal block choice}
\label{sec:theory}
When choosing the temporal block for a ST-GCN model a natural choice would be a recurrent neural network (RNN) (or some subclass of model such as LSTM) due to the time-series nature of the vertex features. RNNs are well suited to supervised learning problems where the dataset has a sequential nature and it has been shown with the right tuning and features they are more flexible and better suited to time-series forecasting than the linear models usually applied \cite{RNN_time_series}. When applied to large datasets (which need not mean large individual time-series but rather many related time-series from the same domain) RNNs have been shown to be industry leading, winning the recent M4 forecasting competition \cite{RNN_time_series_science, RNN_M4}. As the graph features in our prediction task are in the form (even after being convolved by a CNN or GCN) of multiple related time-series, one per vertex, we hypothesise that RNNs should boost prediction performance as they have been shown to learn from multiple time-series simultaneously \cite{RNN_time_series_science}. Simeunović et al. showed that LSTM based ST-GCNs outperform state-of-the-art methods for PV power forecasting while Khodayar et al. showed they also handle noise and uncertainties in time-series data, beating recent state-of-the-art Deep Learning architectures such as deep belief networks and stacked auto-encoders \cite{stgcn_wind}.

Rather than an RNN in the original ST-GCN model Yu et al. opted to use a temporal gated 1-D CNN with a gated linear unit (GLU) as their non-linear activation function \cite{Yu_2018}. They argue despite RNN models being widespread in time-series analysis, they suffer from time-consuming iterations, complex gate mechanisms, and slow response to dynamic changes \cite{Yu_2018}. By contrast, CNNs have the superiority of fast training, simple structures, and no dependency constraints to previous steps \cite{Yu_2018}. The use of a 1-D CNN (as shown in Figure~\ref{fig:framework} with a kernel size of 1xkx1) also allows for sharing of parameters across the vertices even at the temporal level, while still generating distinct vertex features to later be passed into the GCN. Theoretically this allows for a meaningful feature set (corresponding to the CNN ouptut channels) to be produced. We believe the complete removal of CNNs from a ST-GCN model would lead to a performance drop off as these multiple convolved chanels can no longer be fed into the GCN. While Yu et al. make valid arguments in favour of CNNs we hypothesise that, despite being theoretically simpler than LSTMs, CNNs may firstly be over-complicating with their multiple channel outputs ($C_h=32$) and secondly due to the convolutions in the GCN too, CNN based ST-GCNs may struggle with overfitting, despite techniques such as early stopping/dropout. We test for overfitting in Section~\hyperref[sec:experiments]{4} and also look at training times as it has been shown that CNNs perform comparably to LSTMs while being quicker to train \cite{CNN_LSTM}. In other time-series frameworks, such as NLP or stock price prediction, CNNs and LSTMs are both cutting edge and shown to perform with similar accuracy \cite{CNN_RNN_NLP, Mehtab_2020}. Due to how close both architectures perform we hypothesise the two in combination may lead to improved results.

\section{Experiments}
\label{sec:experiments}
We implement Yu et al.'s ST-GCN as our primary model, (referred to as "ST-GCN" from now on). It consists of two stacked CNN-GCN-CNN models and then a linear layer with K $=1$ in both GCNs, GLU activations for the CNNs and ReLU activations for all other layers \cite{Yu_2018}. We also isolate a "CNN-GCN-CNN" model, this time with K $=2$ and everything else the same. This allows us to benchmark how well stacking two 1-step neighbourhoods compares to a direct 2-step neighbourhood. Next we introduce the "GCN-LSTM" model, which is the one used by Simeunović et al., opting for an LSTM temporal block rather than CNN, again with ReLU activations and a final linear layer \cite{stgcn_power_forecasting}. To compare how a combination of temporal block types works we introduce the "CNN-GCN-CNN-LSTM" and "CNN-GCN-LSTM" models, both of which to our knowledge are novel architecture combinations. To make the comparisons fair in our models we keep K $=2$, use GLU activations for the CNN blocks, ReLU activations elsewhere and add a final linear layer. Across all models we use batch normalisation and fix $C_h = 32$ for all CNN models, otherwise $C_h=$ dataset window size.

We use the PyTGT package for our experiments, adapting the stgcn class code \cite{rozemberczki2021pytorch}. We test on $5$ datasets; our finance dataset, a Hungarian chickenpox dataset, a PedalMe bike delivery dataset, a mathematics Wikipedia articles dataset and a Montevideo city bus passengers dataset \cite{rozemberczki2021pytorch}. Each dataset has a rolling window size of either $4$ or $8$. To directly compare models across varying window size we introduce $3$ versions of the Finance dataset, with rolling window sizes $4$, $8$ and $16$.

\begin{table}
\caption{Table of the experiment reuslts for each of the five proposed models across the available datasets. Each cell in the table is of the form: training MSE\,/ testing MSE\,/ training time (minutes) and the dataset information is of the form: (number of vertices, window size, number of samples).}
\label{tab:results}
\centering
\begin{tabular}{llllll}
\toprule
Dataset & ST-GCN & CNN-GCN- & CNN-GCN- & CNN-GCN- & GCN-\\
 & & CNN & CNN-LSTM & LSTM & LSTM\\
\midrule
 Finance-4 & $4.84$\,x$10^{-4}$ & $5.36$\,x$10^{-4}$ & $\mathbf{4.72}$\,x$\mathbf{10^{-4}}$ & $4.73$\,x$10^{-4}$ & $4.73$\,x$10^{-4}$\\
$\left(72, 4, 246\right)$ & $\mathbf{9.31}$\,x$\mathbf{10^{-4}}$ & $9.62$\,x$10^{-4}$ & $9.60$\,x$10^{-4}$  & $9.36$\,x$10^{-4}$ & $9.47$\,x$10^{-4}$\\
& $1.48$ & $1.24$ & $1.87$ & $1.50$ & $\mathbf{3.58}$\,x$\mathbf{10^{-1}}$\\
\hdashline
Finance-8 & $4.87$\,x$10^{-4}$ & $7.56$\,x$10^{-4}$ & $\mathbf{4.76}$\,x$\mathbf{10^{-4}}$ & $4.83$\,x$10^{-4}$ & $5.13$\,x$10^{-4}$\\
$\left(72, 8, 242\right)$ & $1.00$\,x$10^{-3}$ & $1.16$\,x$10^{-3}$ & $\mathbf{9.40}$\,x$\mathbf{10^{-4}}$ & $9.58$\,x$10^{-4}$ & $9.67$\,x$10^{-4}$\\
& $1.49$ & $2.61$ & $3.91$ & $3.44$ & $\mathbf{5.83}$\,x$\mathbf{10^{-1}}$\\
\hdashline
Finance-16 & $5.20$\,x$10^{-4}$ & $6.86$\,x$10^{-4}$ & $5.09$\,x$10^{-4}$ & $\mathbf{4.95}$\,x$\mathbf{10^{-4}} $ & $5.06$\,x$10^{-4}$\\
$\left(72, 16, 234\right)$ & $9.75$\,x$10^{-4}$ & $1.15$\,x$10^{-3}$ & $9.85$\,x$10^{-4}$ & $\mathbf{9.57}$\,x$\mathbf{10^{-4}}$ & $9.66$\,x$10^{-4}$\\
& $5.01$ & $4.59$ & $7.35$ & $5.96$ & $\mathbf{7.86}$\,x$\mathbf{10^{-1}}$\\
\hline
Chickenpox & $6.70$\,x$10^{-1}$ & $\mathbf{6.06}$\,x$\mathbf{10^{-1}}$ & $9.41$\,x$10^{-1}$ & $9.34$\,x$10^{-1}$ & $9.78$\,x$10^{-1}$\\
$\left(20, 4, 517\right)$ & $1.68$ & $1.19$ & $1.08$ & $1.07$ & $\mathbf{1.04}$\\
& $1.52$ & $1.01$ & $1.35$ & $1.14$ & $\mathbf{3.56}$\,x$\mathbf{10^{-1}}$\\
\hline
PedalMe & $\mathbf{1.41}$\,x$\mathbf{10^{-1}}$ & $1.49$\,x$10^{-1}$ & $2.05$\,x$10^{-1}$ & $2.08$\,x$10^{-1}$ & $2.11$\,x$10^{-1}$\\
$\left(15, 4, 31\right)$ & $1.21$ & $\mathbf{1.17}$ & $1.22$ & $1.19$ & $1.24$\\
& $9.92$\,x$10^{-2}$ & $6.30$\,x$10^{-2}$ & $7.27$\,x$10^{-2}$ & $5.78$\,x$10^{-2}$ & $\mathbf{2.17}$\,x$\mathbf{10^{-2}}$\\
\hline
Wikipedia & $4.75$\,x$10^{-1}$ & $\mathbf{4.37}$\,x$\mathbf{10^{-1}}$ & $6.73$\,x$10^{-1}$ & $6.71$\,x$10^{-1}$ & $4.41$\,x$10^{-1}$\\
$\left(1068, 8, 723\right)$ & $1.10$ & $1.31$ & $\mathbf{8.09}$\,x$\mathbf{10^{-1}}$ & $8.15\times10^{-1}$ & $9.19$\,x$10^{-1}$\\
 & $36.1$ & $29.9$ & $54.8$ & $44.4$ & $\mathbf{12.1}$\\
\hline
Bus & $9.47$\,x$10^{-1}$ & $\mathbf{9.40}$\,x$\mathbf{10^{-1}}$ & $9.66$\,x$10^{-1}$ & $9.68$\,x$10^{-1}$ & $1.00$\\
$\left(675, 4, 740\right)$ & $1.02$ & $1.02$ & $\mathbf{9.76}$\,x$\mathbf{10^{-1}}$ & $9.77$\,x$10^{-1}$ & $9.97$\,x$10^{-1}$\\
& $12.6$ & $7.89$ & $12.8$ & $10.3$ & $\mathbf{4.23}$\\
\bottomrule
\end{tabular}
\end{table}

Table~\ref{tab:results} shows the results of our experiments. Firstly, note our combined CNN/LSTM temporal architectures obtain the best test scores on $4/7$ of the datasets, providing strong empirical evidence for our hypothesis that combining both temporal block types improves learning. The Wikipedia dataset is much larger than the others and here our two models perform significantly better than the other architectures. The fact ST-GCN does not outperform our model on these $4$ justifies the improved rate is from CNN-LSTM stacking rather than just increasing model size. We see the GCN-LSTM model is significantly faster for training on all datasets while still having competitive test MSE scores, seemingly contradicting previous literature claiming CNNs train faster than LSTMs \cite{CNN_LSTM, Yu_2018}. We hypothesise this apparent contradiction is in fact due to the CNN inclusion leading to $C_h=32$ in comparison to $C_h=4/8/16$ for GCN-LSTM and thus the GCN specific training is faster. In Section~\hyperref[sec:theory]{3} we speculated that CNNs mutliple $C_h$ channels may lead to over-complication. To test if these are the reasons, we ran the CNN models again now with $C_h=16$ and found noticeable speed increases at the expense of test accuracy.

Comparing CNN-GCN-CNN to CNN-GCN-CNN-LSTM in Table~\ref{tab:results} allows us to directly see how the LSTM addition impacts a model. We observe improved test scores for $6/7$ of the datasets, providing empirical evidence that RNN (LSTM) inclusion indeed boosts model performance as we predicted. If we now compare CNN-GCN-LSTM to GCN-LSTM we see that the CNN removal leads to worst test scores on $6/7$ of the datasets, empirically verifying our hypothesis that CNN removal leads to a performance drop off. Comparing performance on our Finance-$4/8/16$ datasets we see little noticeable change, this could imply the models are robust to window size but further experiments are needed. Due to the noise in the finance dataset we were impressed to see our CNN-GCN-LSTM model outperforming the GCN-LSTM which has been shown to be robust to noise \cite{stgcn_wind}. We hypothesise this is due to the multiple channels distilling signals better. Finally, note $\frac{\textrm{test error}}{\textrm{training error}}$ is on average $2.90$, $2.70$, $2.18$, $2.14$ and $2.26$  for the $5$ models in Table~\ref{tab:results} respectively, showing little evidence that CNNs lead to overfitting however LSTM inclusion clearly reduces overfitting. We attribute this to our use of early stopping and Bhattacharya et al.'s results showing ST-GCNs to not overfit but instead learn meaningful features \cite{stgcn_gaits}. We do however observe clear overfitting on the PedalMe dataset for all models which we believe is due to the training data being too small to learn meaningful features.

\section{Conclusion}
We have provided theoretical arguments for and shown that ST-GCNs with both CNN and LSTM temporal blocks in general have improved performance. We have shown such models are robust to overfitting and relative performance increases on large datasets. We have also shown that having multiple CNN produced hidden channels is beneficial for learning, at the expense of training time.

\newpage

\bibliographystyle{plain}
\bibliography{bibliography}

\begin{thebibliography}{10}

\bibitem{stgcn_gaits}
Uttaran Bhattacharya, Trisha Mittal, Rohan Chandra, Tanmay Randhavane, Aniket Bera, and Dinesh Manocha.
\newblock {STEP:} spatial temporal graph convolutional networks for emotion perception from gaits.
\newblock {\em CoRR}, abs/1910.12906, 2019.

\bibitem{fast_local_spectral}
Micha\"{e}l Defferrard, Xavier Bresson, and Pierre Vandergheynst.
\newblock Convolutional neural networks on graphs with fast localized spectral filtering.
\newblock {\em NeurIPS}, 29, 2016.

\bibitem{grl}
William~L. Hamilton.
\newblock Graph representation learning.
\newblock {\em Synthesis Lectures on Artificial Intelligence and Machine Learning}, 14(3):1--159, 2017.

\bibitem{wavelets}
David~K. Hammond, Pierre Vandergheynst, and Rémi Gribonval.
\newblock Wavelets on graphs via spectral graph theory.
\newblock {\em Applied and Computational Harmonic Analysis}, 30(2):129--150, 2011.

\bibitem{cheb_revisited}
Mingguo He, Zhewei Wei, and Ji{-}Rong Wen.
\newblock Convolutional neural networks on graphs with chebyshev approximation, revisited.
\newblock {\em CoRR}, abs/2202.03580, 2022.

\bibitem{RNN_time_series_science}
Hansika Hewamalage, Christoph Bergmeir, and Kasun Bandara.
\newblock Recurrent neural networks for time series forecasting: Current status and future directions.
\newblock {\em International Journal of Forecasting}, 37(1):388--427, 2021.

\bibitem{stgcn_wind}
Mahdi Khodayar and Jianhui Wang.
\newblock Spatio-temporal graph deep neural network for short-term wind speed forecasting.
\newblock {\em IEEE Transactions on Sustainable Energy}, 10(2):670--681, 2019.

\bibitem{DBLP:journals/corr/KipfW16}
Thomas~N. Kipf and Max Welling.
\newblock Semi-supervised classification with graph convolutional networks.
\newblock {\em CoRR}, abs/1609.02907, 2016.

\bibitem{kipf2016variational}
Thomas~N. Kipf and Max Welling.
\newblock Variational graph auto-encoders, 2016.

\bibitem{Mehtab_2020}
Sidra Mehtab and Jaydip Sen.
\newblock Stock price prediction using {CNN} and {LSTM}-based deep learning models.
\newblock In {\em 2020 International Conference on Decision Aid Sciences and Application ({DASA})}. IEEE, 2020.

\bibitem{RNN_time_series}
G{\'{a}}bor Petneh{\'{a}}zi.
\newblock Recurrent neural networks for time series forecasting.
\newblock {\em CoRR}, abs/1901.00069, 2019.

\bibitem{rozemberczki2021pytorch}
Benedek Rozemberczki, Paul Scherer, Yixuan He, George Panagopoulos, Alexander Riedel, Maria Astefanoaei, Oliver Kiss, Ferenc Beres, Guzman Lopez, Nicolas Collignon, and Rik Sarkar.
\newblock {PyTorch Geometric Temporal: Spatiotemporal Signal Processing with Neural Machine Learning Models}.
\newblock In {\em Proceedings of the 30th ACM International Conference on Information and Knowledge Management}, page 4564–4573, 2021.

\bibitem{signal_processing}
David~I Shuman, Sunil~K. Narang, Pascal Frossard, Antonio Ortega, and Pierre Vandergheynst.
\newblock The emerging field of signal processing on graphs: Extending high-dimensional data analysis to networks and other irregular domains.
\newblock {\em IEEE Signal Processing Magazine}, 30(3):83--98, 2013.

\bibitem{stgcn_power_forecasting}
Jelena Simeunović, Baptiste Schubnel, Pierre{-}Jean Alet, and Rafael~E. Carrillo.
\newblock Spatio-temporal graph neural networks for multi-site {PV} power forecasting.
\newblock {\em CoRR}, abs/2107.13875, 2021.

\bibitem{RNN_M4}
Slawek Smyl.
\newblock A hybrid method of exponential smoothing and recurrent neural networks for time series forecasting.
\newblock {\em International Journal of Forecasting}, 36(1):75--85, 2020.
\newblock M4 Competition.

\bibitem{turner2021graph}
Edward Turner.
\newblock Graph auto-encoders for financial clustering, 2021.

\bibitem{CNN_LSTM}
Hans Weytjens and Jochen~De Weerdt.
\newblock Process outcome prediction: {CNN} vs. {LSTM} (with attention).
\newblock {\em CoRR}, abs/2104.06934, 2021.

\bibitem{Yan_2018}
Sijie Yan, Yuanjun Xiong, and Dahua Lin.
\newblock Spatial temporal graph convolutional networks for skeleton-based action recognition.
\newblock {\em CoRR}, abs/1801.07455, 2018.

\bibitem{CNN_RNN_NLP}
Wenpeng Yin, Katharina Kann, Mo~Yu, and Hinrich Sch{\"{u}}tze.
\newblock Comparative study of {CNN} and {RNN} for natural language processing.
\newblock {\em CoRR}, abs/1702.01923, 2017.

\bibitem{Yu_2018}
Bing Yu, Haoteng Yin, and Zhanxing Zhu.
\newblock Spatio-temporal graph convolutional networks: A deep learning framework for traffic forecasting.
\newblock In {\em Proceedings of the Twenty-Seventh International Joint Conference on Artificial Intelligence}. International Joint Conferences on Artificial Intelligence Organization, 2018.

\end{thebibliography}

\end{document}